\documentclass[10pt,twocolumn,letterpaper]{article}

\usepackage{cvpr}              
\usepackage{float}
\usepackage[table,xcdraw]{xcolor}
\usepackage{multirow}
\usepackage{enumitem}
\usepackage{comment}
\usepackage{duckuments}
\usepackage{subcaption}

\definecolor{cvprblue}{rgb}{0.21,0.49,0.74}
\usepackage[pagebackref,breaklinks,colorlinks,citecolor=cvprblue]{hyperref}

\def\paperID{19} 
\def\confName{CVPR}
\def\confYear{2024}

\title{ST-Gait++: Leveraging spatio-temporal convolutions for\\gait-based emotion recognition on videos}
\author{Maria Luísa Lima\\
Voxar Labs, Centro de Informática\\Universidade Federal de Pernambuco\\
Av. Jorn. Aníbal Fernandes, Recife, Brazil\\
{\tt\small mlll@cin.ufpe.br}
\and
Willams de Lima Costa\\
Voxar Labs, Centro de Informática\\Universidade Federal de Pernambuco\\
Av. Jorn. Aníbal Fernandes, Recife, Brazil\\
{\tt\small wlc2@cin.ufpe.br}
\and
Estefania Talavera Martínez\\
University of Twente\\
Drienerlolaan 5, 7522 NB\\Enschede, Netherlands\\
{\tt\small e.talaveramartinez@utwente.nl}
\and
Veronica Teichrieb\\
Voxar Labs, Centro de Informática\\
Universidade Federal de Pernambuco\\
Av. Jorn. Aníbal Fernandes, Recife, Brazil\\
{\tt\small vt@cin.ufpe.br}
}

\begin{document}
 \twocolumn[{%
 \renewcommand\twocolumn[1][]{#1}%
 \maketitle}]
    

\begin{abstract}
Emotion recognition is relevant for human behaviour understanding, where facial expression and speech recognition have been widely explored by the computer vision community. 
Literature in the field of behavioural psychology indicates that gait, described as the way a person walks, is an additional indicator of emotions. 
In this work, we propose a deep framework for emotion recognition through the analysis of gait. More specifically, our model is composed of a sequence of spatial-temporal Graph Convolutional Networks that produce a robust skeleton-based representation for the task of emotion classification. 
We evaluate our proposed framework on the E-Gait dataset, composed of a total of 2177 samples.
The results obtained represent an improvement of $\approx5\%$ in accuracy compared to the state of the art. In addition, during training we observed a faster convergence of our model compared to the state-of-the-art methodologies.
\end{abstract}

\section{Introduction}
\label{sec:intro}

Humans have a natural perception capability that allows us to capture, process, and understand behavioral cues from other people naturally \cite{lhommet201419}. There are several biological triggers inside our brains that plan our interactions with other people based on this perception \cite{nes2023perception}. However, human behavior is a very broad cognitive spectrum with multiple different nuances that can affect this planning procedure. When looking at social interactions, however, emotion is a specific part of behavior that plays a significant role. The ability to perceive and respond to emotional aspects is essential to develop and maintain links with peers in society.

In multiple contexts of applications, we could argue that understanding the emotions of users is also a significant aspect of the development of systems that are more inclusive and fair. These systems could adapt the way they perform according to how they perceive their users. However, to allow these systems to understand the emotions of their user, first, they need to be able to extract and process emotional information from them. Researchers have been studying how humans perceive and process these affective cues for a while, and evidence from the behavioral psychology literature suggests that a significant part of affective information is communicated naturally and intuitively through a medium known as \textit{nonverbal communication} \cite{buck1991motivation, lhommet2014expressing, jacob2016effects}.

Among the vast list of nonverbal communication cues, some studies highlight the importance of bodily expression for emotion recognition. Early studies such as those by \citet{wallbott1986cues} suggested that there are specific body movements that lead to an accurate perception of emotions. Therefore, this strong evidence from the literature on behavioral psychology validates that the body acts as an outlet for the person's emotional state, as well as signals that, by extracting and processing these cues, someone can perceive emotional aspects through observation of certain characteristics. Recent studies have shown significant success in encoding body language to recognize affect in humans through deep learning and computer vision, indicating that body expressions are a significant cue when building affect-aware technology \cite{kleinsmith2012affective, thuseethan2022emosec, costa2022fast, yang2022emotion}.

However, body language, as well as other affective features such as facial expressions, gaze, gestures, or physiological indicators such as cardiac frequency or respiratory rate, share a common limitation related to applications: they require the user to be facing toward a camera so that the affective sources (e.g., face, eyes, arms...) are visible at all times. Although this constraint is not harmful in some scenarios, such as when people are facing a computer or other affective agents (such as social robots), it is also a limiting aspect in ubiquitous applications since the user would not be able to freely interact with the environment. We could, however, communicate to the user this constraint and ask the user to face the system; however, this would alter the user's behavior, changing how they would communicate their emotions and removing all naturality from this interaction, which would now be an artificial interaction.


A possible way to encode body language and still not limit the user is to look at the person's gait to extract affective information. Gait is the description of the way that someone walks, and previous research in behavioral psychology has found that humans are able to identify multiple social aspects through gait-related parameters \cite{montepare1987identification}. \citet{roether2009critical} applied machine learning to a set of recorded gaits and found that some features, such as movement speed and limb flexion, were essential in correlating emotions and gaits. These pieces of evidence suggest that evaluating gait in a spatiotemporal manner can lead to strong classifications of emotion. With this, new applications in multiple scenarios are enabled, for example: we could leverage gait information to monitor freely moving citizens for public safety, where collective negative emotions could indicate a dangerous situation taking place \cite{https://doi.org/10.1348/135532510X512665}; 
wellbeing applications such as urbanism, where collective spaces can be changed according to what people experience in them \cite{viderman2020correction}; or even
healthcare where the psychological monitoring of a patient in internment can prove to be useful not only as an overall better treatment, especially if non-invasive, but also to enable mental health professionals to better identify mental health issues in their patients by having another information source to look into (\citet{9498861}).

Based on this inspiration, we propose an approach for emotion recognition using gait. Through the use of spatiotemporal processing blocks, we overcome limitations present in the current state-of-the-art \cite{bhattacharya2020step}, which possess a limited processing capability in this aspect. We also show that by overcoming this limitation, our results in the proposed quantitative metric are also better than the state-of-the-art. This work's contributions are as follows:

\begin{itemize}
     \item An emotion recognition method through gait and body language, compatible with behavioral psychology studies, with a $\approx5\%$ increase in accuracy relative to the state-of-the-art 
     \item A model that converges 3.63 times faster in training, saving time and computational resources, allowing for faster testing and experimentation.
\end{itemize}
The rest of the paper is organized as follows: Next, in Section \ref{section:soa}, we visit the state of the art. In Section \ref{section:method}, we present the methods utilized and in Section \ref{sec:exp}, we expose our experimental setup. Later, in Section \ref{sec:results}, we present the results and discuss them. Finally, Section \ref{sec:concl} shows our conclusions and present ideas for future research.
\section{Related Works}
\label{section:soa}
Early works for emotion recognition through gaits were based on extracting features from motion capture data, which is acquired by using special clothing with landmark points and a motion capture camera, and using algorithms to calculate similarity indexes with databases. \citet{venture2014recognizing} captured walks from four different performers and applied PCA to verify that these emotions could be distinguished through some affective features.

A significant improvement was proposed by \citet{daoudi2017emotion}, which evaluated this task from a geometric approach. They have represented skeleton joints over time using covariance matrices, which were mapped to the Riemannian manifold of symmetric positive definite matrices. This allowed the authors to exploit multiple geometric properties for emotion classification. However, this approach is limited by how much temporal information can be encoded, imposing a limited sequence modeling.

The natural next step here was, then, related to improving the temporality aspect of these previous approaches. \citet{randhavane2019learning} presented a new approach, now focused on RNNs, thus allowing improved spatiotemporal relationship. They combined affective features, such as the angles between joints and stride length, with deep features that were learned using a Long Short-Term Memory architecture.

However, the advances on Graph Convolutional Networks at that time, especially the proposal of the ST-GCNs by \citet{Yan_Xiong_Lin_2018} allowed for an even more robust way of learning these relationships. \citet{bhattacharya2020step} proposed using such architecture to extract features from videos and classify the emotions in an implicit manner. 

Still, while ST-GCNs provide an effective approach, there are some limitations that this work aims to address. The representational capacity of the base ST-GCN is predefined rather than learned, which was sufficient for its originally intended application of activity recognition. However, for perceiving emotional cues through nonverbal behaviors like gait, these cues are often more subtle than the movements used for activity recognition. Therefore, not learning the topology may limit the ability to capture these subtle movement patterns that are indicative of different emotions.

This is solved by using the ST-GCN++ architecture, which brings the novelty of learning the topology during training. Another limitation is in the temporal aspect, with ST-GCNs having a fixed size temporal kernel, which was rendered adjustable in the ST-GCN++ architecture. Therefore, switching from ST-GCN to ST-GCN++, is very advantageous for the emotion recognition using gaits problem.

\section{Method}
\label{section:method}


Given a video \(V \in \mathcal{R}^{\mathsf{n} \times \mathsf{h} \times \mathsf{w} \times 3}\) with \(\mathsf{n}\) frames, height \(\mathsf{h}\) \and width \(\mathsf{w}\) and a set of emotions  \(\mathbb{K}\), our task is to classify the perceived emotion of a person present in such video by extracting features related to body language and gait. We first extract a set of 3D body keypoints  \(\mathbf{K} \in \mathcal{R}^{16 \times 3}\), in which \(\mathbf{k}_1, \mathbf{k}_2, \ldots, \mathbf{k}_{16}\), each \(\mathbf{k}_i\) represents the location of a body joint in space related to the person in the video.

 \begin{figure*}[h!]
     \centering
\includegraphics[width=0.9\linewidth]{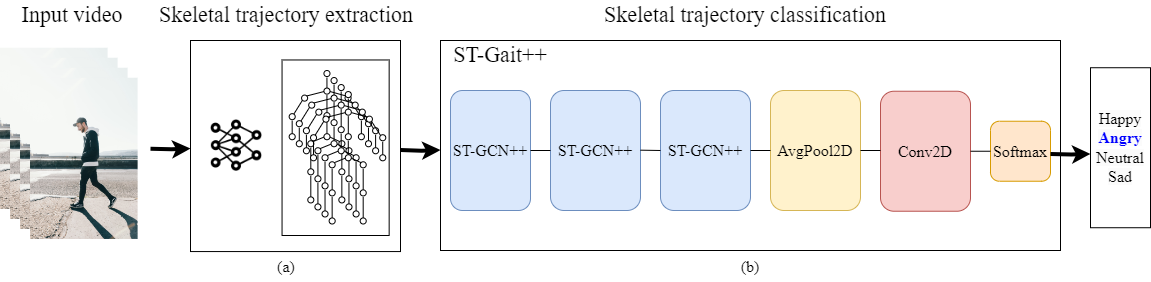}
     \caption{The proposed architecture for ST-Gait++, composed of $3$ ST-GCN++ blocks with outputs sized $32$, $64$ and $64$, followed by a $2D$ Average Pooling and a $1 \times 1$ Convolution, which reduces dimensionality from $64$ to $4$, which is followed by a Softmax. Ideally, in an application scenario, this can be used along some Skeletal trajectory extraction which takes as input a video and outputs the gait sequences to be analysed by ST-Gait++ automatically. This work focuses on the Skeletal trajectory classification.} 
     \label{fig:steppp}
 \end{figure*}


\paragraph{\textbf{(a) Skeletal trajectory extraction.}} One of the possible ways to represent a skeleton is through a graph, since these representations can be considered analog. Each body joint, such as the right shoulder or right elbow can be considered a vertex, and the bones that connects these two joints can be considered edges. This is a clear indicative on why GCNs can be used to process these types of data. Therefore, at a given timestamp \(t\), we extract the skeleton of the person visible on the scene and represent it as a graph: $\mathbb{G} = (\mathbb{V},\mathbb{E})$, where $\mathbb{V}$ is the vertices (or joints) set and $\mathbb{E}$ is the edge (or bones) set and $\mathbb{N}=|\mathbb{V}|$ the number of vertices.
    
    
\paragraph{\textbf{(b) Skeletal trajectory classification.}}

We use this graph \(\mathbb{G}\) as input for our gait processing model. We propose using ST-GCN++ blocks \cite{duan2022pyskl} to learn the joint representations and discover movement patterns related to perceived emotions. This way, nonverbal cues such as step size, arm swinging, head angle relative to the shoulders, among others, can be extracted automatically and without user intervention.

The extracted gait features are propagated from the body joints in the shape of $\mathbb{X} \in \mathcal{R}^{\mathsf{n} \times \mathsf{f}}$ with $x_i \in \mathcal{R}^{\mathsf{f}}$ representing a feature of the $i^{th}$ vertex. The propagation rule is done in the following manner: \(\mathbb{Z}^{(l+1)} = \sigma(\mathbb{AZ}^{(l)}\mathbb{W}^{(l)})\), where $\mathbb{Z}^{l}$ and $\mathbb{Z}^{l+1}$ are the inputs to the network layers $l$ and $l+1$, with $\mathbb{Z}^{0} = \mathbb{X}$. $\mathbb{W}^{l}$ and $\mathbb{W}^{l+1}$ are the weight matrices between layers $l$ and $l+1$, and $\mathbb{A}$ is the adjacency matrix of $\mathbb{G}$ and $\sigma(.)$ is a nonlinear activation function. Each weight matrix $\mathbb{W}$ represents a convolutional kernel which can be used to obtain features. For example, the application of $k$ kernels of dimension $\mathsf{f} \times \mathsf{d}$ in an input $\mathbb{X}$, the output corresponds to a feature vector of dimension $\mathsf{n} \times \mathsf{d} \times \mathsf{k}$. There's also the set of adjacent vertices $\mathit{A}^{t}_{i} \subseteq \mathbb{V}$ to $v^{t}_{i}$ at time $t$ \cite{bhattacharya2020step}.

Our proposed model ST-Gait++ is composed of $3$ ST-GCN++ blocks with $32$, $64$, and $64$ kernels each, followed by an average pooling, a $2$D $1$x$1$ convolution layer, and a softmax layer for the $4$ emotion categories. This design was chosen empirically to overcome the limitations presented previously in Section \ref{section:soa}, but also because other works experiment with similar architectures, such as STEP \cite{bhattacharya2020step}. Also, according to the methodology described by the author \cite{bhattacharya2020step}, affective features can extracted from the data, so those are extracted and added to the used data as well. We overview our model in \autoref{fig:steppp}.

\section{Experimental setup }
\label{sec:exp}
\subsection{Dataset}
We used the Emotion-Gait (E-Gait) dataset in our experiments. The dataset that is currently available is a modified version of the original dataset made available by \citet{bhattacharya2020step}. The authors \cite{bhattacharya2020step} do not share specifics of what has changed or when it did, so we try our best to keep a clear comparison with other techniques from the state of the art. We only used the real data from the E-Gait, because of some quality issues perceived during early experimentation with the synthetic data, alongside the issue of the changes on the dataset.

\begin{figure}[]
     \centering

    \begin{subfigure}[t]{\linewidth}
        \includegraphics[width=\linewidth]{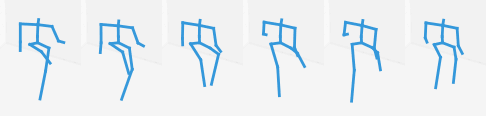}
        \caption{Angry}
    \end{subfigure}
    \begin{subfigure}[t]{\linewidth}
        \includegraphics[width=\linewidth]{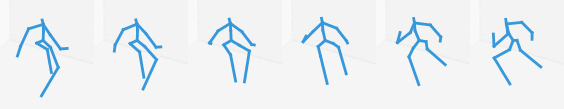}
        \caption{Happy}
    \end{subfigure}
    \begin{subfigure}[t]{\linewidth}
        \includegraphics[width=\linewidth]{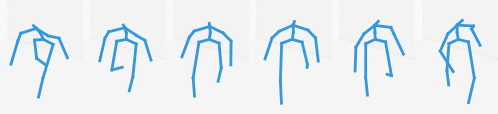}
        \caption{Neutral}
    \end{subfigure}
    \begin{subfigure}[t]{\linewidth}
        \includegraphics[width=\linewidth]{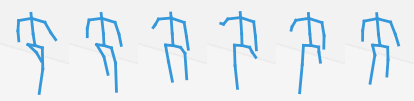}
        \caption{Sad}
    \end{subfigure}
     \caption{Examples of the four categories of the E-Gait dataset. Each item is a sample from one of the categories of the E-Gait dataset, with each frame of the six-frame sequence taken from the whole sample gait sequence. This was done to provide a sense of movement to the reader, so they can better understand E-Gait's characteristics.} 
     \label{fig:dataset-skeletons}
 \end{figure}

The data consists on $342$ samples collected by the authors and 1,835 samples collected by the Edinburgh Locomotion Mocap Database (ELMD) \cite{habibie2017recurrent} and its distribution per train/validation/test split and categories can be found in \autoref{tbl:dataset-dist}. The data is composed of already extracted and normalized skeleton sequences forming different labeled gaits in the $4$ emotions targeted in this paper: Happy, Neutral, Sad and Angry. Each sample is shaped $T \times V$ with $T$ being the number of time steps and $V$ the number of coordinates which is equal to $48$ here since there are $16$ joins with $3$ dimensions each. We show samples of the skeletons present in the dataset in \autoref{fig:dataset-skeletons}.


\begin{table}[b!]
\caption{E-Gait dataset sample distribution according to training, validation and testing splits and the four emotion labels present on E-Gait: Angry, Neutral, Happy and Sad.}
\label{tbl:dataset-dist}
\centering
\resizebox{\columnwidth}{!}{\begin{tabular}{lcccccc}
\hline
                                & \multicolumn{1}{l}{} & \multicolumn{4}{c}{\textbf{Category}}                             & \multicolumn{1}{l}{\multirow{2}{*}{\textbf{Total}}} \\ \cline{3-6}
                                & \multicolumn{1}{l}{} & \textbf{Angry} & \textbf{Neutral} & \textbf{Happy} & \textbf{Sad} & \multicolumn{1}{l}{}                       \\ \cline{2-7} 
\multirow{3}{*}{\textbf{Split}} & \textbf{Train}       & 818            & 332              & 237            & 136          & 1523                                       \\
                                & \textbf{Val}         & 220            & 107              & 65             & 46           & 438                                        \\
                                & \textbf{Test}        & 122            & 48               & 30             & 16           & 216                                        \\ \hline
\end{tabular}}
\end{table}

\subsection{Validation metric}
We have used the \textit{accuracy} metric, which is a common metric used in multiple emotion recognition baselines \cite{costa2023survey}. This metric is also employed in other works that use E-Gait as the evaluation benchmark.


\subsection{Implementation details}

We have used PyTorch $1.7$ to build and train our model. We have used the publicly available implementation for the ST-GCN++ block published by \citet{duan2022pyskl} \footnote{Available at \url{https://github.com/kennymckormick/pyskl}}. From this point, we have applied a Bayesian Search algorithm for hyperparameter tuning. Bayesian Search is a powerful yet simple technique for optimizing hyperparameters by modelling objective functions and updating its belief based on observed results. Therefore, instead of using a random set of hyperparameters, which would be computationally costly, this method will choose the set of hyperparameters that has the highest chance of leading to better results and will discard those with low chance. The search space we navigated using this algorithm is shown in \autoref{tbl:searchSpaceBayes}, and the set of chosen hyperparameters in comparison to those of STEP \cite{bhattacharya2020step} are shown in \autoref{tbl:step-params}. We have trained ST-Gait++ for $200$ epochs.

\begin{table}[H]
\centering
\caption{Search space defined for the Bayesian Search. For each parameter, a set of values are chosen as a search space. As the Bayesian Search finds an optimal value within the search space for each parameter, this value is represented on the third column of the table. These  optimal values were used to train ST-Gait++.}
\label{tbl:searchSpaceBayes}
\resizebox{\columnwidth}{!}{\begin{tabular}{llr}
\hline
\textbf{Parameter}  & \textbf{Search Space} & \textbf{Value} \\ \hline
Basic Learning Rate & 0.01, 0.1, 0.3          & 0.01           \\
GCN Initializer     & Importance, Offset     & Offset         \\
Optimizer           & Adam, RMSProp, SGD      & RMSProp        \\
Weight Decay        & $3 \times 10^{-4}$, $1 \times 10^{-4}$             & $3 \times 10^{-4}$          \\ \hline
\end{tabular}}
\end{table}

\begin{table}[H]
\caption{Training hyperparameters for STEP and ST-Gait++. The learning rate decay happens after epochs 250, 375 and 438 and wasn't used on ST-Gait++. The remaining parameters are the same as STEP, except GCN Initializer which doesn't exist on STEP.}
\label{tbl:step-params}
\centering
\begin{tabular}{lcc}
\hline
\multicolumn{1}{c}{\multirow{2}{*}{\textbf{Parameter}}} & \multicolumn{2}{c}{\textbf{Model}} \\ \cline{2-3} 
\multicolumn{1}{c}{}                                    & \textbf{STEP} & \textbf{ST-Gait++} \\ \hline
Train/Val/Test                                          & 7:2:1         & 7:2:1              \\
Epochs                                                  & 500           & 200                \\
Batch Size                                              & 8             & 8                  \\
Optimizer                                               & Adam          & RMSProp            \\
Basic LR                                                & 0.1           & 0.01               \\
LR Decay                                                & 1/10          & N/A                \\
Momentum                                                & 0.9           & 0.9                \\
Weight Decay                                            & $5\times10^{-4}$          & $3\times10^{-4}$              \\
GCN Initializer                                         & N/A           & Offset             \\ \hline
\end{tabular}
\end{table}

The loss used during training for both STEP and ST-Gait++ was Cross Entropy Loss. We used the Pytorch implementation \footnote{Available at: \url{https://pytorch.org/docs/stable/generated/torch.nn.CrossEntropyLoss.html}}. Since the predictions for each sample were in the form of probabilities for each class, the Cross Entropy Loss can be defined as:

\[l(x,y) = \frac{\sum_{n=1}^{N}(-\sum_{c=1}^{C}w_c log \frac{exp(x_{n,c})}{\sum_{i=1}^{C} exp(x_{n,i})}y_{n,c}
)}{N}\]

Where \(l\) is the loss, \(x\) is the input, \(y\) is the target, \(w\) is the weight \(C\) is the number of classes and \(N\) is the minibatch size.

\section{Results and discussion}
\label{sec:results}



\begin{figure*}[t!]
\centering

\begin{subfigure}[b]{0.45\linewidth}
    \centering
    \includegraphics[width=\textwidth]{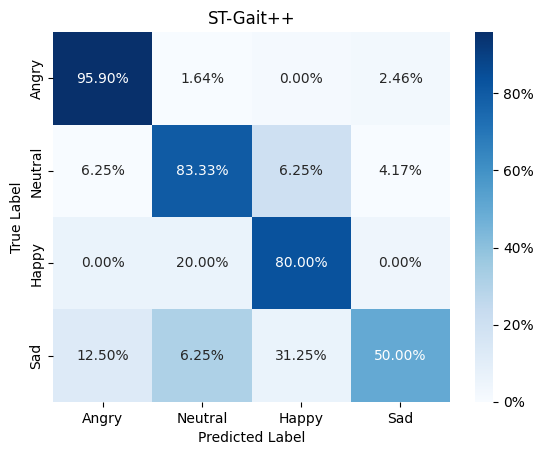}
    \caption{}
\end{subfigure}
\hfill
\begin{subfigure}[b]{0.45\linewidth}
    \centering
    \includegraphics[width=\linewidth]{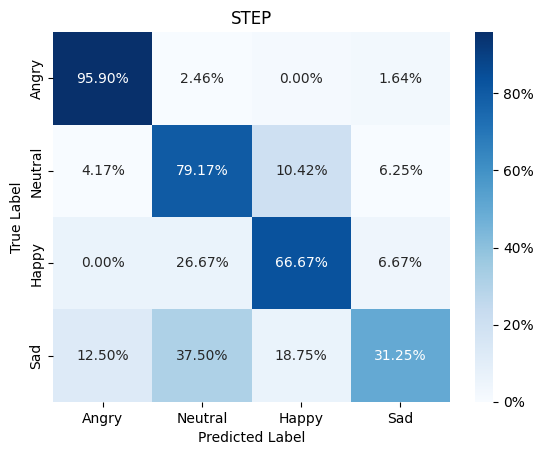}
    \caption{}
\end{subfigure}

\caption{Confusion matrices generated from evaluating the models on the test set for (a) ST-Gait++ and (b) STEP. As can be seen, there is a more pronounced diagonal on (a), emphasizing the better accuracy of ST-Gait++. Also, there is less confusion between the emotions Neutral and Happy on ST-Gait++ than on STEP. However, there is some increase in confusion between Happy and Sad on ST-Gait++.}
\label{img:cm}
\end{figure*}
\paragraph{Quantitative analysis.}

We compare our results with other different approaches in \autoref{tbl:result-comparative}. Our proposed method outperforms other graph-related methods, such as STEP \cite{bhattacharya2020step}. It is interesting to notice that our implementation of this work has led to an increased accuracy than their reported results. In this case, our model has had an increase of $5.4\%$ regarding their result, and $4.2\%$ regarding our implementation. Besides the accuracy increase, our model was also able to converge faster, highlighting several improvements, such as fewer requirements for computational resources or training time, increased possibilities for scaling, and better generalization of data. ST-Gait++ converged on epoch \#127, while STEP converged on epoch \#462, a reduction of $72\%$ in training time. A runtime analysis was also performed, over the test set, and can be found on table \autoref{tbl:runtime}. As observed, STEP has a better time for inference, which can be explained by the simpler architecture.

\begin{table}[]
\caption{Runtime analysis on the test set, which has a total of 216 samples. The test was performed on a laptop with a NVIDIA GeForce RTX 4050 GPU.}
\label{tbl:runtime}
\centering
\resizebox{\columnwidth}{!}{
\begin{tabular}{ccc}
\hline
\textbf{Method}  & \textbf{Min Inference Time} & \textbf{Average Inference Time} \\ \hline
STEP             & 0.99ms ($\sim$1001 fps)     & 3.28ms ($\sim$304 fps)          \\
ST-Gait++ (ours) & 4.64ms ($\sim$215 fps)      & 15.08ms ($\sim$66 fps)          \\ \hline
\end{tabular}}
\end{table}

Our model also highly outperforms other approaches with a temporal focus, such as \citet{randhavane2019identifying}. In this case, we report an accuracy increase of $6.8\%$. Overall, this quantitative evaluation highlights that our model not only has increased accuracy in relation to the current state-of-the-art, but is also more optimized towards training requirements.

\begin{table}[b!]
\caption{Quantitative analysis of methods for emotion recognition using the E-Gait dataset.}
\label{tbl:result-comparative}
\centering
\resizebox{\columnwidth}{!}{\begin{tabular}{
>{\columncolor[HTML]{FFFFFF}}l 
>{\columncolor[HTML]{FFFFFF}}c }
\hline
\textbf{Methods}                                & \textbf{Acc (\%)} \\ \hline
\citeauthor{venture2014recognizing} (\citeyear{venture2014recognizing})  & 30.8             \\
\citeauthor{daoudi2017emotion} (\citeyear{daoudi2017emotion})       & 42.5                   \\
\citeauthor{li2016identifying} (\citeyear{li2016identifying})   & 53.7                   \\
\citeauthor{crenn2016body} (\citeyear{crenn2016body}) & 66.2                   \\
\citeauthor{randhavane2019identifying} (\citeyear{randhavane2019identifying})  & 80.7         \\
\citeauthor{narayanan2020proxemo} (\citeyear{narayanan2020proxemo})                         & 82.4                   \\
\citeauthor{bhattacharya2020step} (\citeyear{bhattacharya2020step}) (STEP)                     & 82.1                   \\
Bhattacharya et al. (2020) (STEP) (Our implementation) & 83.3                   \\ 
\citeauthor{yumeng2024affective} (\citeyear{yumeng2024affective})  & 85.2             \\ \hline
\textbf{ST-Gait++ (Ours)}                       & \textbf{87.5}          \\ \hline
\end{tabular}}
\end{table}

We have also generated confusion matrices for ST-Gait++ and STEP in order to compare how the accuracy of these two models can be represented in how they perceive the emotional classes differently. We show these results in \autoref{img:cm}. As expected, this result reaffirm the results we have discussed before; STEP has a higher confusion between \textit{Neutral} and \textit{Happy} than our model. In the case of facial expressions, the literature on behavioral psychology has shown that there is a structural resemblance between neutral and happy faces that could lead to confusion, and it is common to have this type of ambiguity in these tasks \cite{said2009structural, costa2022fast}; however, there is still no evidence that the same could happen to gait perception. Therefore, it is difficult to judge if STEP's ambiguity between these two classes could be justified. In our model, we also notice an ambiguity between Sad and Happy. We do not have any insights regarding this curious behavior at this point, but we also highlight that our accuracy for the Sad class is still higher than the reported for STEP.

\begin{figure}
    \centering
    \includegraphics[width=\linewidth]{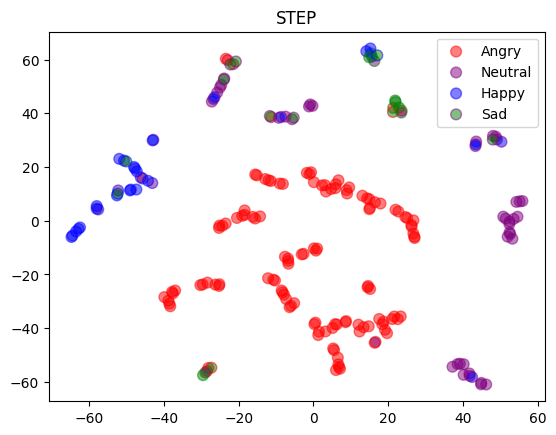}
    \includegraphics[width=\linewidth]{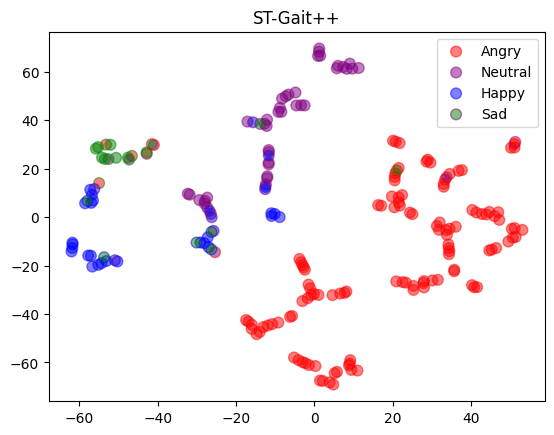}

    \caption{T-SNE representation of the features extracted from the last convolutional layer of ST-Gait++ and STEP on the test set, The inner color represents the ground truth labels and the outer circle represents the model's inference.}
    \label{fig:tsne-stgait}
\end{figure}

Finally, we have also applied a t-SNE representation of the features extracted from the last layer of the models to visualize and analyze the learned representations of the data. The last layer usually contains higher-level, abstract features of the input data, which could usually be represented by activation maps or feature maps. In this case, since we are not directly working with images at that point, applying t-SNE could reveal clusters or groups that could highlight the ability of the model to distinguish classes or categories effectively, as well as to identify outliers or anomalies in the data. We show in \autoref{fig:tsne-stgait} that the Angry class has a very distinct and separated clustering, which explains the better performance in this class as we have shown before in \autoref{img:cm}. This better performance can also be attributed to the far larger amount of samples for this category, making the model better able to separate it from the rest. Neutral and Happy categories are distinguishable enough to show why they also show a good performance, while Sad, the least numerous category, is very mingled with the others. 

\begin{figure}[]
     \centering

    \begin{subfigure}[t]{\linewidth}
        \includegraphics[width=\linewidth]{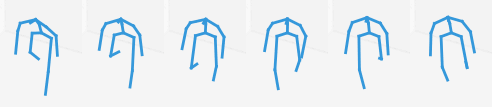}
        \caption{Ground Truth: Happy | ST-Gait++ Prediction: Happy}
    \end{subfigure}
         \begin{subfigure}[t]{\linewidth}
        \includegraphics[width=\linewidth]{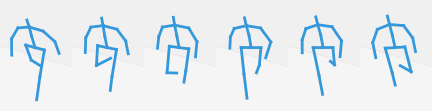}
        \caption{Ground Truth: Sad | ST-Gait++ Prediction: Sad}
    \end{subfigure}
    \begin{subfigure}[t]{\linewidth}
        \includegraphics[width=\linewidth]{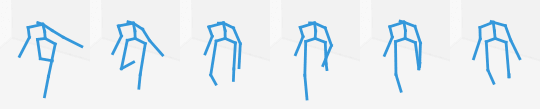}
        \caption{Ground Truth: Angry | ST-Gait++ Prediction: Sad}
    \end{subfigure}
     \begin{subfigure}[t]{\linewidth}
        \includegraphics[width=\linewidth]{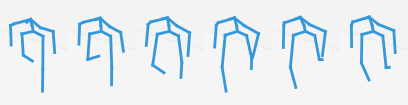}
        \caption{Ground Truth: Neutral | ST-Gait++ Prediction: Happy}
    \end{subfigure}
    \caption{Examples of correct and incorrect inferences by ST-Gait++ on the test set.}
    \label{fig:stgait-mistakes}
\end{figure}

\paragraph{Qualitative analysis.} In \autoref{fig:stgait-mistakes} we show some qualitative visualizations with examples of correct and incorrect predictions. By looking just at the skeletons it's impossible to distinguish the emotions. For some, we may agree with the annotation provided and understand why the model made a correct prediction. For others, it's dubious to infer the perceived emotion with just the skeleton, and we can understand why the model made a mistake. On (a) we can see the wide stride, arm swinging and attribute that to happiness, but we can also see the somewhat lowered head, which could indicate sadness. On (b) we can se a fast stride and arm swinging that could indicate some more energetic emotion, but the model still correctly classifies it as sadness. (c) gives us a lowered head, which could have lead the model to infer on sadness, even though the swinging arms and big stride indicate anger. Lastly, (d) shows a fast stride with slightly swinging arms and upperbody which could have misled the model into inferring happiness and not some neutral emotion. Overall, we can understand that there is some dubiousness in the data that leads to explainable mistakes. 

However, given the temporality aspect of this evaluation, we are limited to what we can show in this research paper. We have prepared a video with a more in-depth overview available at to be made available upon publication.


\paragraph{Limitations of the E-Gait dataset.} To the best of our knowledge, E-Gait is one of the only public datasets of emotion recognition through gait, alongside Emotion-Walk (E-Walk) \cite{randhavane2019identifying}. However, since they have data overlap, we decided to test on E-Gait, given that it is the one used by the state-of-the-art approaches of emotion recognition from gait. 

In addition, the E-Gait dataset \cite{bhattacharya2020step} is very obscure, with only the skeletons being available to researchers. Because of this, the annotations cannot be confirmed and neither can we know the transformations taken to process the videos and the skeletons which are already normalized. This can be observed in \autoref{fig:stgait-mistakes}. Furthermore, the dataset does not disclose the demographics of the subjects and a great part of the data comes from the capture with a single individual and adds a heavy bias. For an application in other contexts, such as Latin American contexts, for example, it would be very interesting to have a dataset that could show the local cultural emotion expression. Also, publishing an open dataset also containing the original videos, not only the skeletons, would give researchers a higher freedom for experimenting, validating ideas and checking for biases to correct.

The first step to making such a dataset would be to gather other available datasets for emotion recognition through gait, or curate new ones, focusing on bringing high quality data that is representative, diverse and as unbiased as possible. Testing ST-Gait++ on more data will certainly point to new paths of improvement for this research.

\paragraph{Diversity and Bias in emotion recognition related datasets.}
There are many studies focused on trying to find biases in intelligent systems. One such study is \citet{rhue2018racial}, which found racial disparities in facial emotion recognition and raised the question of whether artificial inteligence could in fact determine emotion better than people. To answer this question, we need to understand that demographics such as gender, race and ethnicity heavily influence the perception of human characteristics, such as in facial emotion recognition algorithms \cite{buolamwini2017gender}. But this is also due to the fact that we as humans perceive the world based on our own biases. To annotate the data that will be fed into algorithms is to accept that the data will have biases that the model will propagate. Studies such as  \citet{pahl2022female} bring to attention the age bias, besides the ethnicity bias, in prominent Action Unit Datasets. The algorithms studied had problems with glasses but not with beards. This shows data collection problem: Are there no diverse people available or are these people not even considered as a possible variation in the target user public? These limitations can be extended to gaits, as it is a characteristic and emotion expression outlet that can vary across different demographics.

Furthermore, although some emotion expression may have universal features \cite{tracy2008spontaneous}, it is noted that cultural particularities are very influential in affective cues encoding \citet{kleinsmith2012affective}. For a general application, having an analysis on the demographic characteristics can help researchers gain more insight on the limitations of the data and, as a consequence, of the real world application of their research. As such, having such a skewed dataset, such as ELMD (\citet{habibie2017recurrent}), being a considerable part of the total data available, brings to question the applicability of the entire dataset in in-the-wild scenarios. 

Given the ELMD dataset, of the $342$ samples recorded, 90 participants were involved, with no demographic of this public being disclosed. It is disclosed that the data, all $1,835$ samples, was collected using a sole male actor. With this in mind, it's important to point out the importance of diversity and fairness in this data collection. 

Also, it's important to emphasize that what is being dealt with here are the perceived emotions, since whatever actual emotion was being felt by the person at the time of data capture is only available if the person is questioned (in real scenarios) or if we know what they are trying to convey (actors portraying emotions). Even in real scenarios, the reported emotion after questioning could be biased due to the type of question asked or recall biases \cite{hernandez2021guidelines}.

\section{Conclusion}
\label{sec:concl}

In this work, we proposed ST-Gait++, a novel framework for emotion recognition from gait. Its skeleton-based spatio-temporal representation approach results in state-of-the-art classification performance on the E-Gait dataset. We also discussed some of the limitations of the field with the objective of presenting research opportunities.

In addition, given the faster training convergence on a consumer grade laptop, we expect ST-Gait++ to provide new research opportunities in the field of human behaviour analysis for researchers with a lower budget or limited resources.

Future work will explore the relevance of the different body parts for recognising emotion and the inclusion of additional gait descriptors.

{
    \small
    \bibliographystyle{ieeenat_fullname}
    \bibliography{main}
}


\end{document}